\documentclass[runningheads]{llncs}
\usepackage{graphicx}
\usepackage{inconsolata}
\usepackage{graphicx}
\usepackage{amsmath}
\usepackage{bm}

\usepackage{amsthm}
\usepackage{amssymb}
\usepackage{multirow}
\usepackage{booktabs}
\usepackage{color}
\usepackage{caption}
\usepackage{subcaption}
\usepackage[colorlinks,
            linkcolor=blue,
            anchorcolor=blue,
            citecolor=blue,
            urlcolor=blue
           ]{hyperref}
\usepackage{marvosym}
\usepackage{ifsym}

\makeatletter
\def\thanks#1{\protected@xdef\@thanks{\@thanks
        \protect\footnotetext{#1}}}
\makeatother

\begin{document}
\title{Exploiting Pseudo Future Contexts for Emotion Recognition in Conversations\thanks{This work was supported by the National Key R\&D Program of China [2022YFF0902703].}}
\titlerunning{ERCMC}
%
\author{Yinyi Wei\inst{1} \and
Shuaipeng Liu\inst{2}\textsuperscript{\Letter} \and
Hailei Yan\inst{2} \and \\
Wei Ye\inst{3} \and
Tong Mo\inst{1} \and
Guanglu Wan\inst{2}}
\authorrunning{Yinyi Wei et al.}
%
\institute{Peking University
\and Meituan Group, Beijing, China
\and National Engineering Research Center for Software Engineering, Peking University \\
\email{wyyy@pku.edu.cn, liushuaipeng@meituan.com}}

%
%
\maketitle              
\begin{abstract}
With the extensive accumulation of conversational data on the Internet, emotion recognition in conversations (ERC) has received increasing attention. Previous efforts of this task mainly focus on leveraging contextual and speaker-specific features, or integrating heterogeneous external commonsense knowledge. Among them, some heavily rely on future contexts, which, however, are not always available in real-life scenarios. This fact inspires us to generate pseudo future contexts to improve ERC. Specifically, for an utterance, we generate its future context with pre-trained language models, potentially containing extra beneficial knowledge in a conversational form homogeneous with the historical ones. These characteristics make pseudo future contexts easily fused with historical contexts and historical speaker-specific contexts, yielding a conceptually simple framework systematically integrating multi-contexts. Experimental results on four ERC datasets demonstrate our method's superiority. Further in-depth analyses reveal that pseudo future contexts can rival real ones to some extent, especially in relatively context-independent conversations.
\end{abstract}

\section{Introduction}
Emotion recognition in conversations (ERC), aiming to identify the emotion of each utterance in a conversation, is an essential part of conversation understanding. It plays a significant role for various downstream tasks, e.g., opinion mining in social media, building intelligent assistant through empathetic machine, health care, and detecting fake news in social media \cite{lin2019moel,shu2022cross}.

\begin{figure}[t]
    \centering
    \includegraphics[width=0.6\linewidth]{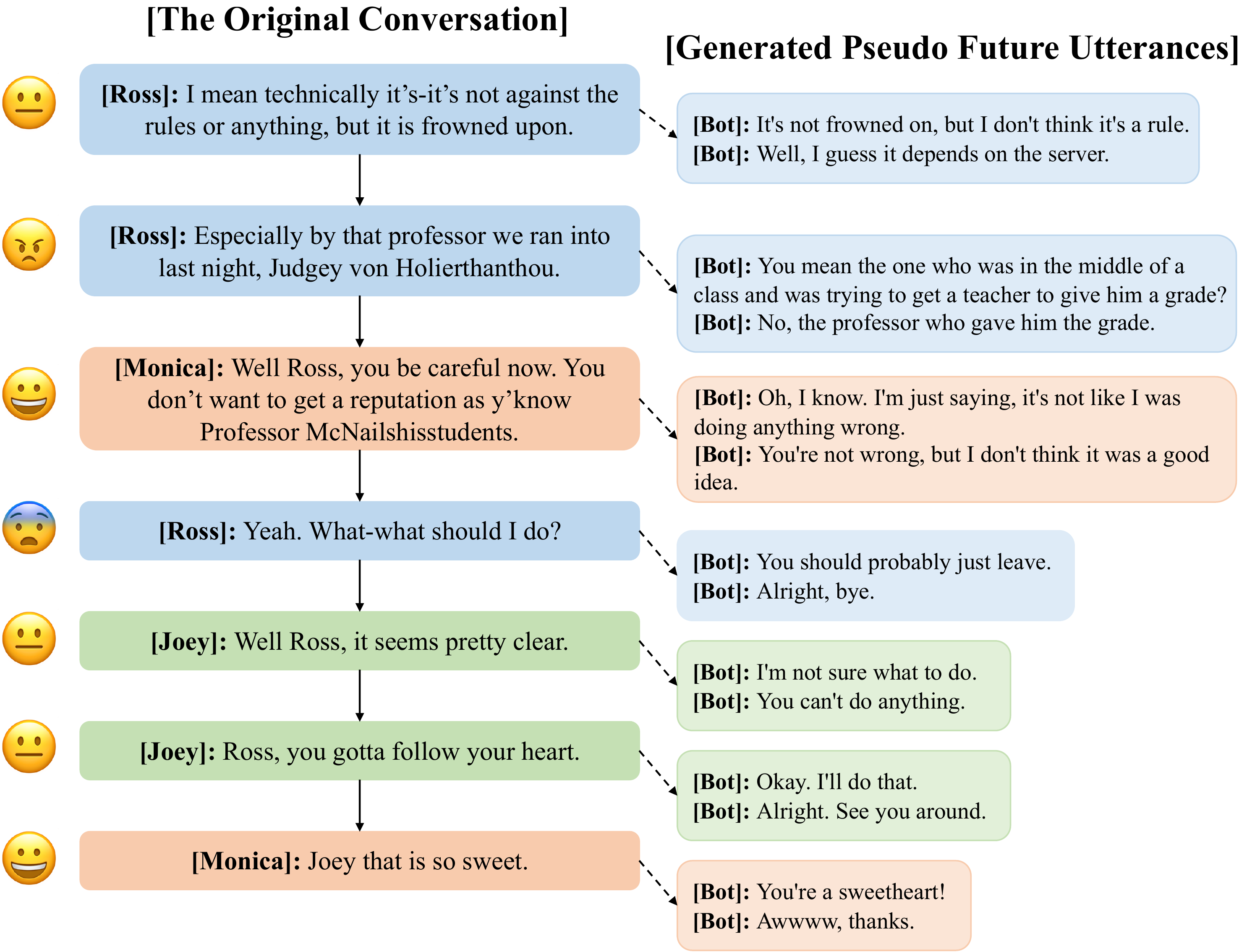}
    \caption{An conversation of MELD. Original utterances are in dark-colored blocks and generated pseudo future contexts are in light-colored blocks.}
    \label{fig1}
\end{figure}

A common solution for ERC is to jointly leverage the contextual and speaker information of an utterance, which can be roughly divided into sequence-based methods \cite{majumder2019dialoguernn,li2022contrast} and graph-based methods \cite{ghosal2019dialoguegcn,shen2021directed}. To further enrich conversational information, another line of research proposes to enhance representations of utterances with knowledge augmentation. These methods are mainly devoted to introducing external commonsense knowledge to facilitate comprehension of utterances, and have amply proved their effectiveness \cite{ghosal2020cosmic,zhu2021topic}.

Though many sophisticated designs have been proposed to exploit context information, we found some of them heavily rely on future contexts \cite{ghosal2019dialoguegcn}. The problem is that we may face the unavailability of future contexts. For example, imaging a scenario of detecting the emotion of an utterance just-in-time in an after-sale conversation, we will have no future contexts available. This fact motivates us to simulate unseen future states by generating pseudo future contexts. As shown in Figure~\ref{fig1}, for each utterance, we can use a pre-trained language model (e.g., DialoGPT \cite{zhang2020dialogpt}) to generate its future context, introducing consistent yet extra beneficial knowledge for emotion recognition. Compared with heterogeneous external knowledge used in previous works (e.g., commonsense knowledge), the knowledge extracted is represented in a conversational form homogeneous with historical contexts, and hence can be easily integrated.

We further design a novel context representation mechanism that can be applied indiscriminately to multi-contexts, including historical contexts, historical speaker-specific contexts, and pseudo future contexts. Specifically, for each type of contexts, we employ relative position embeddings to capture the positional relationship between an utterance and its surroundings, obtaining a refined representation of an utterance, as well as a local conversational state. We then utilize GRUs to model how the local state evolves as the conversation progresses, characterizing long-term utterance dependencies from a global view. A final representation, which fuses multi-contexts information from both local and global perspectives, is fed into a classifier to produce the emotion label of an utterance. 

We conduct extensive experiments on four ERC datasets, and the empirical results verify the effectiveness and potential of our method.

In summary, our contributions are: (1) We propose improving ERC from a novel perspective of generating pseudo future contexts for utterances, incorporating utterance-consistent yet potentially diversified knowledge from pre-trained language models. (2) We design a simple yet effective ERC framework to integrate pseudo future contexts with historical ones and historical speaker-specific ones, yielding competitive results on four widely-used ERC datasets. (3) We analyze how pseudo future contexts correlate with characteristics of emotion-consistency and context-dependency among utterances in conversations, revealing that pseudo future contexts resemble real ones in some scenarios.

\section{Related Works}
Works on ERC can be roughly categorized into sequence-based methods and graph-based methods.

(1) \textbf{Sequence-Based Methods}.
Sequence-based methods treat each utterance as a discrete sequence with RNNs \cite{elman1990finding} or Transformers \cite{vaswani2017attention}. \cite{majumder2019dialoguernn} and \cite{hu2021dialoguecrn} utilized RNNs to track context and speaker states. \cite{li2020hitrans} used two levels of Transformers to capture contextual information and seize speaker information with an auxiliary task. \cite{wang2020contextualized} treated ERC as an utterance-level sequence tagging task with a conditional random field layer. \cite{li2022contrast} utilized BART with supervised contrastive learning and an auxiliary response generation task.

(2) \textbf{Graph-Based Methods}.
Treating an utterance as a node, contextual and speaker relations as edges, ERC can be modelled using graph neural networks. \cite{ghosal2019dialoguegcn} modelled both the context- and speaker-sensitive dependencies with graph convolutional networks. To reflect the relational structure, \cite{ishiwatari2020relation} proposed relational position encodings in graph attention networks. Considering the temporal property, \cite{shen2021directed} proposed to encode with a directed acyclic graph.

(3) \textbf{ERC with External Knowledge}.
Some works propose to introduce heterogeneous commonsense knowledge for ERC. \cite{zhong2019knowledge} combined word and concept embeddings with two knowledge sources. Taking advantage of COMET, a Transformer-based generative model \cite{bosselut2019comet} for generating commonsense descriptions, \cite{ghosal2020cosmic,zhu2021topic,li2021past} enhanced the representations of utterances with commonsense knowledge. It is noteworthy that our method distinguishes itself by not relying on the design of intricate instructions or prompts to extract knowledge from the pre-trained language models, setting it apart from other methods that utilize generative models as knowledge bases \cite{wei2022eliciting,ling2023evolutionary}.

\section{Methodology}
\begin{figure}[t]
    \centering
    \includegraphics[width=0.9\linewidth]{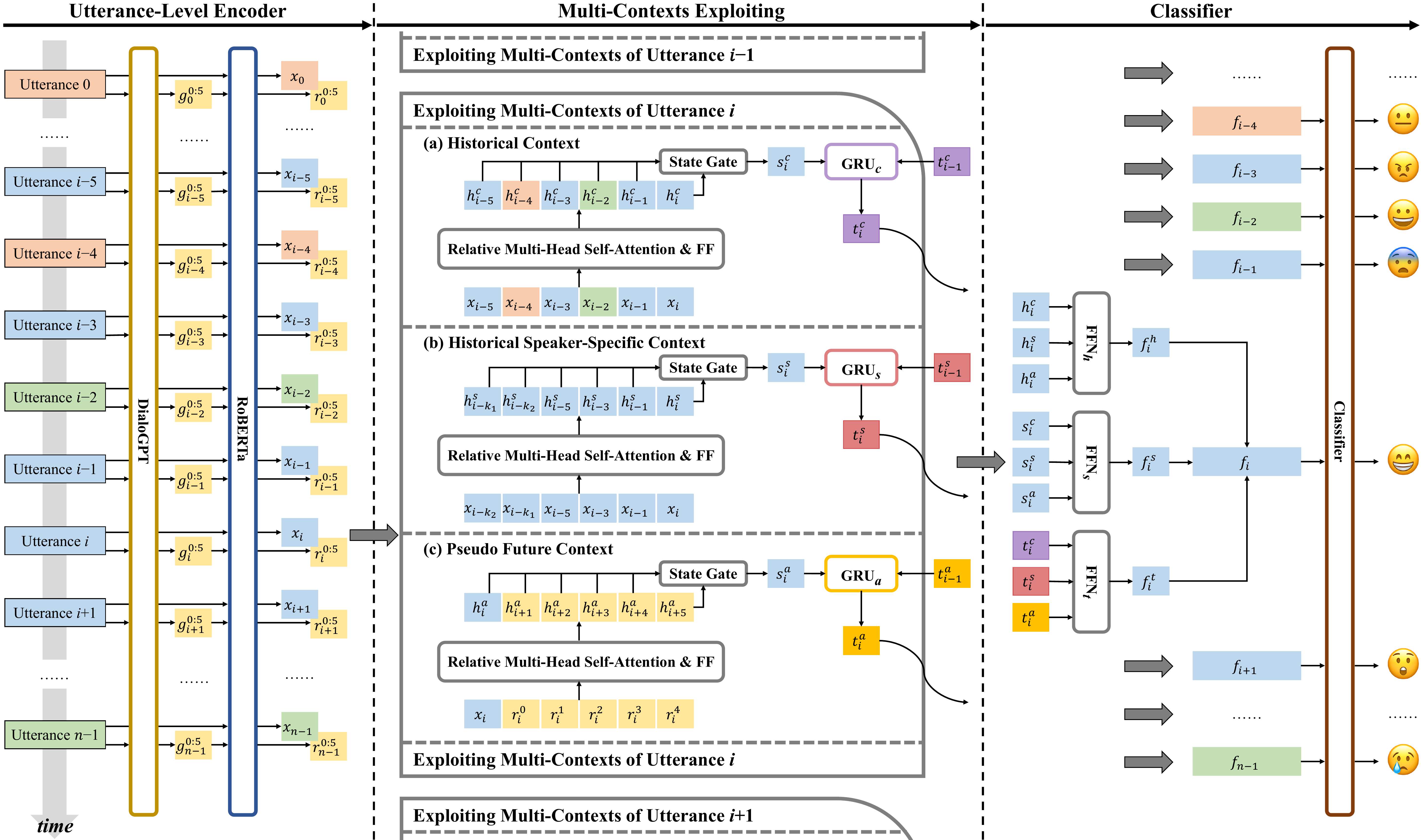}
    \caption{The overall architecture of our proposed ERCMC framework. Utterances from various speakers are in different coloured blocks (e.g., orange, blue and green). Generated utterances are in yellow blocks.}
    \label{fig2}
\end{figure}
\subsection{Task Definition}
Formally, denote $\mathcal{U}$, $\mathcal{P}$ and $\mathcal{Y}$ as conversation set, speaker set and label set. For a conversation $U\in\mathcal{U}$, $U=(u_0, \cdots, u_{n-1})$, where $u_i$ is the $i$-th utterance. The speaker of $u_i$ is denoted by function $P(\cdot)$. For example, $P(u_i)=p_j, p_j\in\mathcal{P}$ means that $u_i$ is uttered by $p_j$. The goal of ERC is to assign an emotion label $y_i\in\mathcal{Y}$ to each $u_i$, formulated as an utterance-level sequence tagging task in this work. As with most previous works \cite{ghosal2020cosmic,zhu2021topic}, for an utterance, we only use its historical utterances, while future utterances are not available to fit real-life scenarios.

\subsection{Generation of Pseudo Future Contexts}
\label{ka}
Knowing that DialoGPT \cite{zhang2020dialogpt} is trained on large-scale conversational data (e.g., a 27GB Reddit dataset) and can generate responses of high quality, we employ it as a knowledge base to introduce homogeneous external knowledge.

Given an utterance $u_i$, we utilize a DialoGPT $\mathcal{G}$ to recursively generate a pseudo future context $g_i^{0:m}$ by considering at most $k$ historical utterances prior to $u_i$, where $k=\max{(i, k)}$, $m$ is the maximum number of generated utterances.
\begin{align}
    g_i^0 &= \mathcal{G}(u_{i-k},\cdots,u_i) \\
    g_i^j &= \mathcal{G}(u_{i-k},\cdots,u_i,g_i^0,\cdots,g_i^{j-1})\;, \ j\in [1,m)
\end{align}

\subsection{Proposed Framework}
Figure~\ref{fig2} shows the overall architecture, which is named ERCMC (\textbf{E}motion \textbf{R}ecognition in \textbf{C}onversations with \textbf{M}ulti-\textbf{C}ontexts).

\subsubsection{Utterance-Level Encoder}
Given a conversation $U=(u_0, \cdots,u_{n-1})$, for $u_i\in U$, we first obtain $m$ generated pseudo future utterances $g_i^{0:m}$. Then the utterance and its pseudo future utterances are fed into an encoder $\mathcal{M}$ (RoBERTa \cite{liu2019roberta} in our experiments) to obtain their representations (we represent an utterance with the embedding of the ${\tt{[CLS]}}$ token).
\begin{align}
    {x}_i &= \mathcal{M}(u_i) \\
    {r}_i^j &= \mathcal{M}(g_i^j)
\end{align}
where ${x}_i, {r}_i^j\in\mathbb{R}^{d_{m}}$. ${x}_i$ and ${r}_i^j$ are the representations of $u_i$ and $g_i^j$ respectively.

\subsubsection{Multi-Contexts Exploiting}
To exploit multi-contexts for an utterance, we consider the three local areas for ${x}_i$: the historical context $C_i$, the historical context of the same speaker $S_i$ and the future pseudo context $A_i$.
\begin{align}
    &C_i = ({x}_{i-\ell}, \cdots, {x}_{i-1}, {x}_i) \\
    &S_i = ({x}_{b_\ell}, \cdots, {x}_{b_1}, {x}_i) \\
    &A_i = ({x}_i, {r}_i^0, \cdots, {r}_i^{m-1})
\end{align}
where $\ell+1$ is the size of the three local areas. In $S_i$, $P(u_{b_j})=P(u_i)$. In $C_i$ and $S_i$, $\ell=\max{(i, \ell)}$, and in $A_i$, $m$ is set to the initial value of $\ell$.

For each local area of an utterance, we calculate three representations focusing on local-aware state, local state, and evolution of local states. We take $C_i$ as an example in the following calculation.

We first use a multi-head self-attention layer with relative position embeddings \cite{shaw2018self} to obtain a local-aware embedding considering $C_i$:
\begin{align}
    &h_j^c = \max{(0,{h'}_j^cW^{PF}_1+b^{PF}_1)W^{PF}_2} + b^{PF}_2 \\
    &{h'}_j^c = (head_{1j}^c \mathbin\Vert \cdots \mathbin\Vert head_{n_hj}^c) W^O \\
    &head^c_{pj} = \sum_{k=i-\ell}^{i}\alpha_{jk}(x_kW_p^V+rp_{pjk}^V) \\
    &\alpha_{jk} = \frac{\exp{e_{jk}}}{ {\textstyle \sum_{o=i-\ell}^{i}}\exp{e_{jo}}} \\
    &e_{jk}=\frac{x_jW^Q_p(x_kW_p^K+rp^K_{pjk})^\top}{\sqrt{d_k}} 
\end{align}
where $h_j^c\in\mathbb{R}^{d_{m}}$, $\mathbin\Vert$ is the concatenation of two vectors, $W^{PF}_1\in\mathbb{R}^{d_m\times 4d_m},b^{PF}_1\in\mathbb{R}^{4d_m},W^{PF}_2\in\mathbb{R}^{4d_m\times d_m},b^{PF}_1\in\mathbb{R}^{d_m}$. The number of head is $n_h$, $W^O\in\mathbb{R}^{d_vn_h\times d_m}$. For each head $p$, $W^Q_p,W^K_p\in \mathbb{R}^{d_m\times d_k}$, $W^V_p\in \mathbb{R}^{d_m\times d_v}$ are projections, $rp_{pjk}^V\in\mathbb{R}^{d_v}$ and $rp^K_{pjk}\in\mathbb{R}^{d_k}$ are relative position embeddings between $x_j$ and $x_k$. Similarly, for $S_i$ and $A_i$, two embeddings $h^s$ and $h^a$ are obtained.

A state gate is then used to get the local state $s_i^c$:
\begin{align}
    &s_i^c = \sum_{j=i-\ell}^{i-1}\beta_j h_j^c \\
    &\beta_j=\frac{\exp{st_j}}{ {\textstyle \sum_{k=i-\ell}^{i-1}} \exp{st_k}} \\
    &st_j = \tanh{(h^c_jW^S(h^c_i)^\top)}
\end{align}
where $s_i^c\in\mathbb{R}^{d_{m}}$, $W^S\in \mathbb{R}^{d_m\times d_m}$. Similarly, for $S_i$ and $A_i$, two local states $s_i^s$ and $s_i^a$ are obtained.

For states in a global view, a GRU unit is utilized to characterize the evolution of local states:
\begin{align}
    t_i^c = {\rm{GRU}}(s_i^c, t_{i-1}^c)
\end{align}
where $t^c_i\in\mathbb{R}^{d_{m}}$ is the tracked global state prior to $u_i$ and $t^c_0$ is initialized with zero. Similarly, for $S_i$ and $A_i$, two GRU units are used to get $t^s_i$ and $t^a_i$.

\subsubsection{Classifier}
For $u_i$, the exploited outcomes from multi-contexts are integrated into a final representation $f_i$.
\begin{align}
    &f_i = f^h_i \mathbin\Vert f^s_i \mathbin\Vert f^t_i \\
    &f^h_i = (h^c_i \mathbin\Vert h^s_i \mathbin\Vert h^a_i)W^F_h \\
    &f^s_i = (s^c_i \mathbin\Vert s^s_i \mathbin\Vert s^a_i)W^F_s \\
    &f^t_i = (t^c_i \mathbin\Vert t^s_i \mathbin\Vert t^a_i)W^F_t
\end{align}
where $W^F_h,W^F_s,W^F_t\in\mathbb{R}^{3d_{m}\times d_m}$.

To obtain the labels $Y=(y_0, \cdots,y_{n-1})$ with the representations $F=(f_0, \cdots,f_{n-1})$, we apply a feed forward network and a softmax layer:
\begin{align}
    &y_i = \textrm{argmax}(P_i) \\
    &P_i = \textrm{softmax}(F_iW^M+b^M)
\end{align}
where $W^M\in\mathbb{R}^{3d_m\times|\mathcal{Y}|}$ and $b^M\in\mathbb{R}^{|\mathcal{Y}|}$.

The model is trained using negative log-likelihood loss.

\section{Experimental Setups}
\subsection{Datasets and Evaluation Metrics}
\begin{table}[t]
\centering
\caption{Statistics of datasets.}
\begin{tabular}{l|ccc|ccc|c}
\toprule
Dataset     & \multicolumn{3}{c|}{Conversations}                               & \multicolumn{3}{c|}{Utterances}                                  & \multirow{2}{*}{Classes} \\ \cmidrule{2-7}
            & \multicolumn{1}{c|}{Train}  & \multicolumn{1}{c|}{Dev}   & Test  & \multicolumn{1}{c|}{Train}  & \multicolumn{1}{c|}{Dev}   & Test  &                          \\ \midrule
IEMOCAP     & \multicolumn{2}{c|}{120}                                 & 31    & \multicolumn{2}{c|}{5,810}                               & 1,623 & 6                        \\ \midrule
DailyDialog & \multicolumn{1}{c|}{11,118} & \multicolumn{1}{c|}{1,000} & 1,000 & \multicolumn{1}{c|}{87,170} & \multicolumn{1}{c|}{8,069} & 7,740 & 7                        \\ \midrule
EmoryNLP    & \multicolumn{1}{c|}{659}    & \multicolumn{1}{c|}{89}    & 79    & \multicolumn{1}{c|}{7,551}  & \multicolumn{1}{c|}{954}   & 984   & 7                        \\ \midrule
MELD        & \multicolumn{1}{c|}{1,038}  & \multicolumn{1}{c|}{114}   & 280   & \multicolumn{1}{c|}{9,989}  & \multicolumn{1}{c|}{1,109} & 2,610 & 7                        \\ \bottomrule
\end{tabular}
\label{table1}
\end{table}

We evaluate our proposed framework on four ERC datasets: IEMOCAP \cite{busso2008iemocap}, DailyDialog \cite{li2017dailydialog}, EmoryNLP \cite{zahiri2018emotion}, MELD \cite{poria2019meld}. IEMOCAP and DailyDialog are two-party datasets, while EmoryNLP and MELD are multi-party datasets. For the four datasets, we only use the textual parts. Statistics of these datasets are shown in Table~\ref{table1}.

Following \cite{ghosal2020cosmic,zhu2021topic}, we choose weighted-average F1 for IEMOCAP, EmoryNLP and MELD. Since the \emph{neutral} class constitutes to 83\% of the DailyDialog, micro-averaged F1 excluding \emph{neutral} is chosen.

\subsection{Baselines}
For models without external knowledge, we compare with: (1) \textbf{Sequence-based methods}: \textbf{DialogueRNN} \cite{majumder2019dialoguernn} uses GRUs to track context and speaker states; \textbf{HiTrans} \cite{li2020hitrans} utilizes two level Transformers with an auxiliary task to leverage both contextual and speaker information; \textbf{CoG-BART} \cite{li2022contrast} uses BART with supervised contrastive learning and a response generation task. (2) \textbf{Graph-based Methods}: \textbf{DialogueGCN} \cite{ghosal2019dialoguegcn} models dependencies about context and speaker with graph convolutional networks; \textbf{RGAT} \cite{ishiwatari2020relation} uses relational graph attention networks with relational position encodings.

For models with external knowledge, we compare with: \textbf{KET} \cite{zhong2019knowledge} uses a Transformer to combine word and concept embeddings; \textbf{COSMIC} \cite{ghosal2020cosmic} is a modified DialogueRNN with commonsense knowledge from COMET; \textbf{TODKAT} leverages COMET to integrate commonsense knowledge and a topic model to detect the potential topics of a conversation; \textbf{SKAIG} \cite{li2021past} models structural psychological interactions between utterances with commonsense knowledge.

Three variants of ERCMC are compared: (1) ERCMC without future contexts; (2) ERCMC with multi-contexts; (3) ERCMC using real future contexts. 

\subsection{Implementation Details}

\begin{table}[t]
\centering
\caption{Overall results. In each part, the highest scores are in boldface. * indicates using future contexts. C, S, PF, and RF denote historical contexts, historical speaker-specific contexts, pseudo future contexts, and real future contexts.}
\begin{tabular}{llcccc}
\toprule
\multicolumn{2}{l|}{\multirow{2}{*}{Methods}}                                     & \multicolumn{1}{c|}{IEMOCAP}          & \multicolumn{1}{c|}{DailyDialog}   & \multicolumn{1}{c|}{EmoryNLP}         & MELD                            \\ \cmidrule{3-6} 
\multicolumn{2}{l|}{}                                                             & \multicolumn{1}{c|}{Weighted F1} & \multicolumn{1}{c|}{Micro F1} & \multicolumn{1}{c|}{Weighted F1} & Weighted F1                \\ \midrule
\multicolumn{6}{c}{Without External Knowledge}                                                                                                                                              \\ \midrule
\multicolumn{2}{l|}{DialogueRNN} & \multicolumn{1}{c|}{62.57}   & \multicolumn{1}{c|}{55.95}       & \multicolumn{1}{c|}{31.70}    & 57.03 \\
\multicolumn{2}{l|}{\quad + RoBERTa}                                                      & \multicolumn{1}{c|}{64.76}   & \multicolumn{1}{c|}{57.32}       & \multicolumn{1}{c|}{37.44}    & 63.61 \\
\multicolumn{2}{l|}{DialogueGCN*}                                                 & \multicolumn{1}{c|}{64.18}   & \multicolumn{1}{c|}{-}           & \multicolumn{1}{c|}{-}        & 58.10 \\
\multicolumn{2}{l|}{\quad + RoBERTa*}                                                     & \multicolumn{1}{c|}{64.91}   & \multicolumn{1}{c|}{57.52}       & \multicolumn{1}{c|}{38.10}    & 63.02 \\
\multicolumn{2}{l|}{RGAT*}                                                        & \multicolumn{1}{c|}{65.22}   & \multicolumn{1}{c|}{54.31}       & \multicolumn{1}{c|}{34.42}    & 60.91 \\
\multicolumn{2}{l|}{\quad +RoBERTa*}                               & \multicolumn{1}{c|}{\textbf{66.36}}   & \multicolumn{1}{c|}{\textbf{59.02}}       & \multicolumn{1}{c|}{37.89}    & 62.80 \\
\multicolumn{2}{l|}{HiTrans*}                                                     & \multicolumn{1}{c|}{64.50}   & \multicolumn{1}{c|}{-}           & \multicolumn{1}{c|}{36.75}    & 61.94 \\
\multicolumn{2}{l|}{CoG-BART*}                                                    & \multicolumn{1}{c|}{66.18}   & \multicolumn{1}{c|}{56.29}       & \multicolumn{1}{c|}{\textbf{39.04}}    & \textbf{64.81} \\ \midrule
\multicolumn{6}{c}{With External Knowledge}                                                                                                                                                 \\ \midrule
\multicolumn{2}{l|}{KET}                                                          & \multicolumn{1}{c|}{59.56}   & \multicolumn{1}{c|}{53.37}       & \multicolumn{1}{c|}{34.39}    & 58.18 \\
\multicolumn{2}{l|}{COSMIC}                                                       & \multicolumn{1}{c|}{65.28}   & \multicolumn{1}{c|}{58.48}       & \multicolumn{1}{c|}{38.11}    & 65.21 \\
\multicolumn{2}{l|}{SKAIG*}                                                       & \multicolumn{1}{c|}{\textbf{66.96}}   & \multicolumn{1}{c|}{\textbf{59.75}}       & \multicolumn{1}{c|}{\textbf{38.88}}    & 65.18 \\
\multicolumn{2}{l|}{TODKAT}                                                       & \multicolumn{1}{c|}{61.33}   & \multicolumn{1}{c|}{58.47}       & \multicolumn{1}{c|}{38.69}    & \textbf{65.47} \\ \midrule
\multicolumn{6}{c}{Variants of Our Model}                                                                                                                                                   \\ \midrule
\multicolumn{1}{l|}{\multirow{3}{*}{ERCMC}}     & \multicolumn{1}{l|}{C \& S}        & \multicolumn{1}{c|}{65.47}   & \multicolumn{1}{c|}{59.85}       & \multicolumn{1}{c|}{38.71}    & 65.21 \\
\multicolumn{1}{l|}{}                        & \multicolumn{1}{l|}{C \& S \& PF}  & \multicolumn{1}{c|}{66.07}   & \multicolumn{1}{c|}{59.92}       & \multicolumn{1}{c|}{\textbf{39.34}}    & \textbf{65.64} \\
\multicolumn{1}{l|}{}                        & \multicolumn{1}{l|}{C \& S \& RF*}  & \multicolumn{1}{c|}{\textbf{66.51}}   & \multicolumn{1}{c|}{\textbf{61.33}}       & \multicolumn{1}{c|}{38.90}    & 65.43 \\ \bottomrule
\end{tabular}
\label{table2}
\end{table}

We use DialoGPT-medium to generate 5 pseudo future utterances for each utterance (i.e., $m=5$) with $k$ historical utterances, where $k=2$ for IEMOCAP and DailyDialog, and $k=4$ for EmoryNLP and MELD. For backbones, we select RoBERTa-base for IEMOCAP and RoBERTa-large for the other datasets. For each utterance, the max length is set to 128. For multi-contexts exploiting, we set the size of the three local areas to 6 (i.e., $\ell = 5$). The number of head $n_h$ is set to 8. $d_m$ is set to 768 for IEMOCAP and 1,024 for the other datasets. $d_k,d_v=\frac{d_m}{n_h}$. For training setup, the batch size is set to 1 conversation for IEMOCAP and DailyDialog, and 2 conversations for EmoryNLP and MELD. The gradient accumulation step is set to 16 for DailyDialog and 4 for IEMOCAP, EmoryNLP and MELD. We train the model for 20 epochs for IEMOCAP and 10 epochs for the other datasets with the AdamW optimizer whose learning rate is set to 3e-5 for IEMOCAP and 1e-5 for the other datasets. Each model is trained for max epochs and choose the checkpoint with the best validation performance. The dropout rate is set to 0.1. All the experiments are done on a single NVIDIA Tesla V100. All of our results are the average of 5 runs. We do not implement on larger generative models, such as ChatGPT, due to their significant computational resources requirements and associated costs. Our code is available at \url{https://github.com/Ydongd/ERCMC}.

\section{Results and Insights}
\subsection{Overall Results}
\label{overall}
Overall results are shown in Table~\ref{table2}, from which we have three main observations: (1) Compared with models using future contexts, ERCMC in C\&S\&RF setting achieves competitive performance, demonstrating the superiority of our proposed framework. (2) In stead of leveraging heterogeneous commonsense knowledge as previous works, we use homogeneous conversational knowledge, allowing ERCMC in C\&S\&PF setting to overtake other models using external knowledge. (3) Additional future contexts provide more information about conversations and thus bring improvements over C\&S setting when using pseudo or real future contexts in our model. Compared with using real future contexts, using pseudo future contexts performs better on EmoryNLP and MELD, and underperforms on IEMOCAP and DailyDialog. An in-depth analysis is given in Section~\ref{future}.

\subsection{Collaboration of Multi-Contexts}
\label{collaboration}

\begin{table}[t]
\centering
\caption{Various combinations of Multi-Contexts. RAW denotes no contex.}
\begin{tabular}{lcccc}
\toprule
Part        & IEMOCAP & DailyDialog & EmoryNLP & MELD  \\
\midrule
RAW     & 56.48   & 57.46       & 37.78    & 64.06 \\
\midrule
C           & 63.95   & 59.14       & 37.88    & 64.20 \\
S           & 64.39   & 59.48       & 37.97    & 64.43 \\
PF           & 57.38   & 58.16       & 37.84    & 64.20 \\
\midrule
C \& PF      & 62.29   & 59.50       & 37.90    & 64.36 \\
S \& PF      & 63.35   & 59.66       & 37.98    & 64.76 \\
C \& S      & 65.47   & 59.85       & 38.71    & 65.21 \\
\midrule
C \& S \& PF & 66.07   & 59.92       & 39.34    & 65.64 \\
\bottomrule
\end{tabular}
\label{table3}
\end{table}

To reveal the collaborative effect of multi-contexts on ERCMC, we show various combinations of multi-contexts in Table~\ref{table3}, from which we observe that: (1) All combinations improve the performance compared with only using raw information, suggesting that the context information is critical for ERC. (2) Regardless of the number of contexts used, settings with historical speaker-specific contexts produce the best results, indicating that historical contexts uttered by the same speaker is the predominant part. (3) Using C or S alone, and a combination of both, bring more improvements than using PF and its combination. The observation implies that historical contexts and historical speaker-specific contexts are more significant for ERC, while pseudo future contexts serve as a supplementary role to provide extra knowledge.

\subsection{Ablation Study}

\subsubsection{Effect of Compositions of the Final Representations.}
\label{compositions}

\begin{table}[t]
\caption{Ablation study of ERCMC in C\&S\&PF setting.}
\begin{subtable}{0.48\textwidth}
\centering
\caption{Results with different compositions of the final representations, $h$, $s$, and $t$ denote local-aware embedding, local state, and tracked global state, respectively.}
\begin{tabular}{lcccc}
\toprule
Dataset     & w/o $h$ & w/o $s$ & w/o $t$ & $h$, $s$, $t$ \\ \midrule
IEMOCAP     & 62.88   & 64.35   & 64.81   & 66.07         \\
DailyDialog & 59.16   & 59.59   & 59.50   & 59.92         \\
EmoryNLP    & 20.08   & 38.65   & 38.74   & 39.34         \\
MELD        & 51.51   & 65.06   & 65.30   & 65.64         \\ \bottomrule
\end{tabular}
\label{table4}
\end{subtable}
\begin{subtable}{0.48\textwidth}
\centering
\caption{Results with different position embeddings. N, S, L, and R denote using no embeddings, sinusoidal, learnable, and relative positon embeddings, respectively.}
\begin{tabular}{lcccc}
\toprule
Dataset     & N                         & S                         & L                         & R                       \\ \midrule
IEMOCAP     & 65.48                     & 64.61                     & 65.28                     & 66.07                     \\
DailyDialog & 59.89 & 59.90 & 59.83 & 59.92 \\
EmoryNLP    & 38.64 & 38.51 & 38.57 & 39.34 \\
MELD        & 64.98                     & 64.83                     & 64.41                     & 65.64                     \\ \bottomrule
\end{tabular}
\label{table5}
\end{subtable}
\end{table}

We form the final representations with local-aware embeddings, local states and tracked global states, which are denoted as $h$, $s$, and $t$, respectively. Results with removal of each part are shown in Table~\ref{table4}, from which we can find that though the calculation of local states and tracked global states contain some information from local-aware embeddings, performances drop dramatically when original embeddings are lost, especially on EmoryNLP and MELD. And performances fall slightly when local states and tracked global states are removed.

\subsubsection{Effect of Position Embeddings.}
Unlike modeling the distances between utterances with sinusoidal position embeddings in previous sequence-based methods \cite{zhong2019knowledge,li2020hitrans}, we utilize relative position embeddings and further compare with three position embeddings in Table~\ref{table5}: without position embeddings; sinusoidal position embeddings that use sine and cosine functions of different frequencies; position embeddings learned from scratch. It is clear that the distance information exploited by relative position embeddings at utterance-level is more compatible with a sequence-based method and therefore achieves better results.

\subsubsection{Effect of Size of Local Areas.}
\label{size}

\begin{figure}[t]
    \centering
    \includegraphics[width=0.6\linewidth]{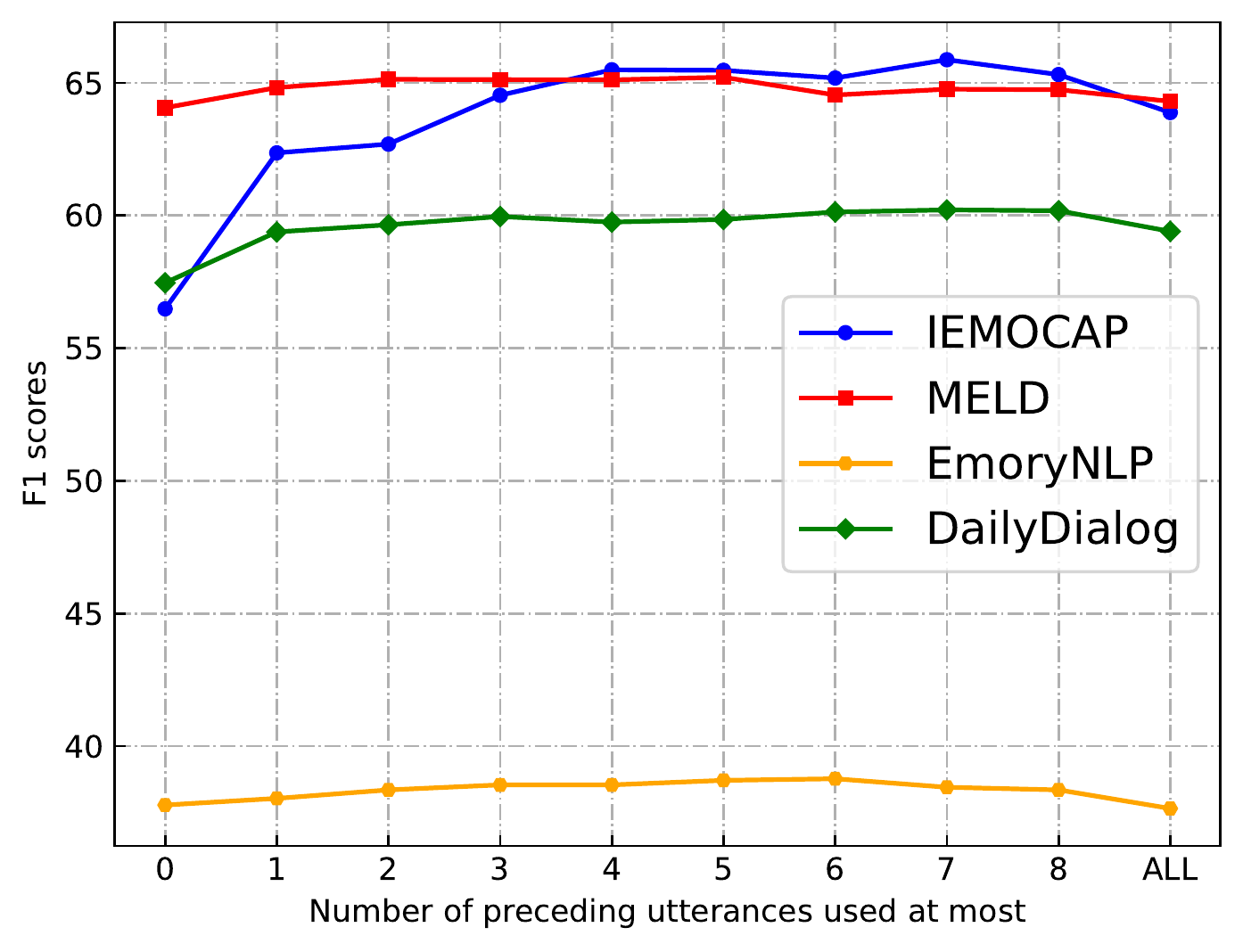}
    \caption{Effect of number of historical utterances with ERCMC in C\&S setting.}
    \label{fig3}
\end{figure}

In our experiments, we set the size of local areas to 6 (i.e., using five historical utterances at most). We further investigate the size effect of local areas by employing ERCMC in C\&S setting. From the results in Figure~\ref{fig3}, we can draw a general trend that performances show an upward tendency when starting using historical utterances and decrease when using all historical contexts due to information redundancy. We can also see that compared with IEMOCAP and DailyDialog, MELD and EmoryNLP have a smoother performance, especially at the beginning and end of the curve.

\subsection{Future Context: Pseudo or Real}
\label{future}

\begin{table}[t]
\caption{Performance and emotion-consistency on four simplified test sets.}
\begin{subtable}{0.5\textwidth}
\centering
\caption{Simplified test set of IEMOCAP with 1468 utterances.}
\begin{tabular}{l|ccc}
\toprule
\multirow{2}{*}{Setting} & \multicolumn{3}{c}{IEMOCAP}                                   \\ \cmidrule{2-4} 
                         & Performance & $WT_1$                    & $WT_2$                    \\ \midrule
PF                       & 57.81       & \multirow{2}{*}{35.85} & \multirow{2}{*}{38.18} \\
C \& S \& PF             & 66.30       &                        &                        \\ \midrule
RF                       & 62.81       & \multirow{2}{*}{50.10} & \multirow{2}{*}{50.47} \\
C \& S \& RF             & 66.68       &                        &                        \\ \bottomrule
\end{tabular}
\label{table6}
\end{subtable}
\begin{subtable}{0.5\textwidth}
\centering
\caption{Simplified test set of DailyDialog with 3123 utterances.}
\begin{tabular}{l|ccc}
\toprule
\multirow{2}{*}{Setting} & \multicolumn{3}{c}{DailyDialog}             \\ \cmidrule{2-4} 
                         & Performance & $WT_1$                    & $WT_2$                    \\ \midrule
PF                       & 51.19       & \multirow{2}{*}{44.77} & \multirow{2}{*}{60.43} \\
C \& S \& PF             & 53.80       &                        &                        \\ \midrule
RF                       & 53.78       & \multirow{2}{*}{76.79} & \multirow{2}{*}{78.90} \\
C \& S \& RF             & 54.53       &                        &                        \\ \bottomrule
\end{tabular}
\label{table7}
\end{subtable}
\begin{subtable}{0.5\textwidth}
\centering
\caption{Simplified test set of EmoryNLP with 608 utterances.}
\begin{tabular}{l|ccc}
\toprule
\multirow{2}{*}{Setting} & \multicolumn{3}{c}{EmoryNLP}                 \\ \cmidrule{2-4} 
                         & Performance & $WT_1$                    & $WT_2$                    \\ \midrule
PF                       & 40.94       & \multirow{2}{*}{29.95} & \multirow{2}{*}{31.36} \\
C \& S \& PF             & 41.86       &                        &                        \\ \midrule
RF                       & 40.64       & \multirow{2}{*}{27.11} & \multirow{2}{*}{29.22} \\
C \& S \& RF             & 41.73       &                        &                        \\ \bottomrule
\end{tabular}
\label{table8}
\end{subtable}
\begin{subtable}{0.5\textwidth}
\centering
\caption{Simplified test set of MELD with 1360 utterances.}
\begin{tabular}{l|ccc}
\toprule
\multirow{2}{*}{Setting} & \multicolumn{3}{c}{MELD}                    \\ \cmidrule{2-4} 
                         & Performance & WT1                    & WT2                    \\ \midrule
PF                       & 64.07       & \multirow{2}{*}{39.52} & \multirow{2}{*}{43.00} \\
C \& S \& PF             & 65.68       &                        &                        \\ \midrule
RF                       & 63.69       & \multirow{2}{*}{35.62} & \multirow{2}{*}{38.38} \\
C \& S \& RF             & 64.97       &                        &                        \\ \bottomrule
\end{tabular}
\label{table9}
\end{subtable}
\end{table}

In Section~\ref{overall}, we have mentioned that both pseudo and real future contexts improve the performance of ERCMC, and compared with the C\&S\&RF setting, the C\&S\&PF setting performs better on EmoryNLP and MELD and worse on IEMOCAP and DailyDialog. This section provides a more in-depth analysis of how pseudo and real future contexts affect the final results.

We first categorize the four datasets concerning several previous observations.

Firstly, in Table~\ref{table3}, C\&S setting obtains relative improvements over RAW setting of 15.92\% on IEMOCAP, 4.16\% on DailyDialog, 2.46\% on EmoryNLP and 1.80\% on MELD. Secondly, in Table~\ref{table4}, compared with the full C\&S\&PF setting, when the original local-aware embeddings are removed, the performances drops by 4.83\% on IEMOCAP, 1.27\% on DailyDialog, 48.96\% on EmoryNLP and 21.53\% on MELD. Thirdly, in Figure~\ref{fig3}, the curves of EmoryNLP and MELD are stabler than those of IEMOCAP and DailyDialog.

From them, we can conclude that conversations in IEMOCAP and DailyDialog are more context-dependent, while conversations in EmoryNLP and MELD are relatively context-independent. An alternative explanation for this conclusion is that IEMOCAP and DailyDialog are two-party datasets, and therefore the conversations are more emotion-focused, whereas EmoryNLP and MELD are multi-party datasets, resulting in more diffuse conversations and inconsistent emotions.

To further explore the effect of using pseudo or real future contexts, we attempt to investigate the attributes of future contexts. Previous works \cite{ghosal2021exploring,yang2022hybrid} have reported a common issue with ERC: different emotions in consecutive utterances may confound a model and thereby degrade performance. Inspired by the issue, we define a new concept of ``emotion-consistency'' as the degree of emotional consistency of the subsequent utterances with the first utterance within a local area. Given a local area $LC=(u_0, \cdots, u_\ell)$, the emotion-consistency is:
\begin{align}
    \textrm{EC}(LC)=100\cdot\sum_{i=1}^{\ell} \phi(u_i,u_0)\cdot wt_{i-1}
\end{align}
where $WT=(wt_0, \cdots, wt_{\ell-1})$ is the weight in different positions, $\phi(u_i,u_0)$ is a function to indicate whether $u_i$ and $u_0$ having the same emotion.

For quantitative analysis, we employ ERCMC in C setting to predict emotions and calculate the emotion-consistency scores. To align with pseudo future contexts, we simplify the test sets to ensure each utterance has at least $\ell$ consecutive utterances ($\ell=5$ in our experiments). Two kinds of weight are considered, $wt_i^1=\frac{1}{\ell}, wt_i^1\in WT_1; wt_i^2=\frac{\exp{e^{\ell-i}}}{ {\textstyle \sum_{j=1}^{\ell}}\exp{e^{j-1}} }, wt_i^2\in WT_2$. $WT_1$ relies uniformly on consecutive utterances, while $WT_2$ favours utterances near the first utterance.

From Table~\ref{table6}, \ref{table7}, \ref{table8} and \ref{table9}, we can observe that: (1) Emotion-consistency scores of IEMOCAP and DailyDialog are higher than those of EmoryNLP and MELD. The finding confirms our previous observations that IEMOCAP and DailyDialog are more context-dependent, whereas EmoryNLP and MELD are somewhat context-independent. (2) Performance and emotion-consistency scores are positively correlated. And combined with Table~\ref{table2}, using pseudo future contexts achieves competitive results on IEMOCAP and DailyDialog, and outperforms using real future contexts on EmoryNLP and MELD. The results demonstrate that pseudo future contexts can replace real ones to some extent when the dataset is context-dependent, and serve as more extra beneficial knowledge when the dataset is relatively context-independent.

\subsection{Case Study}

\begin{figure}[t]
    \centering
    \includegraphics[width=0.65\linewidth]{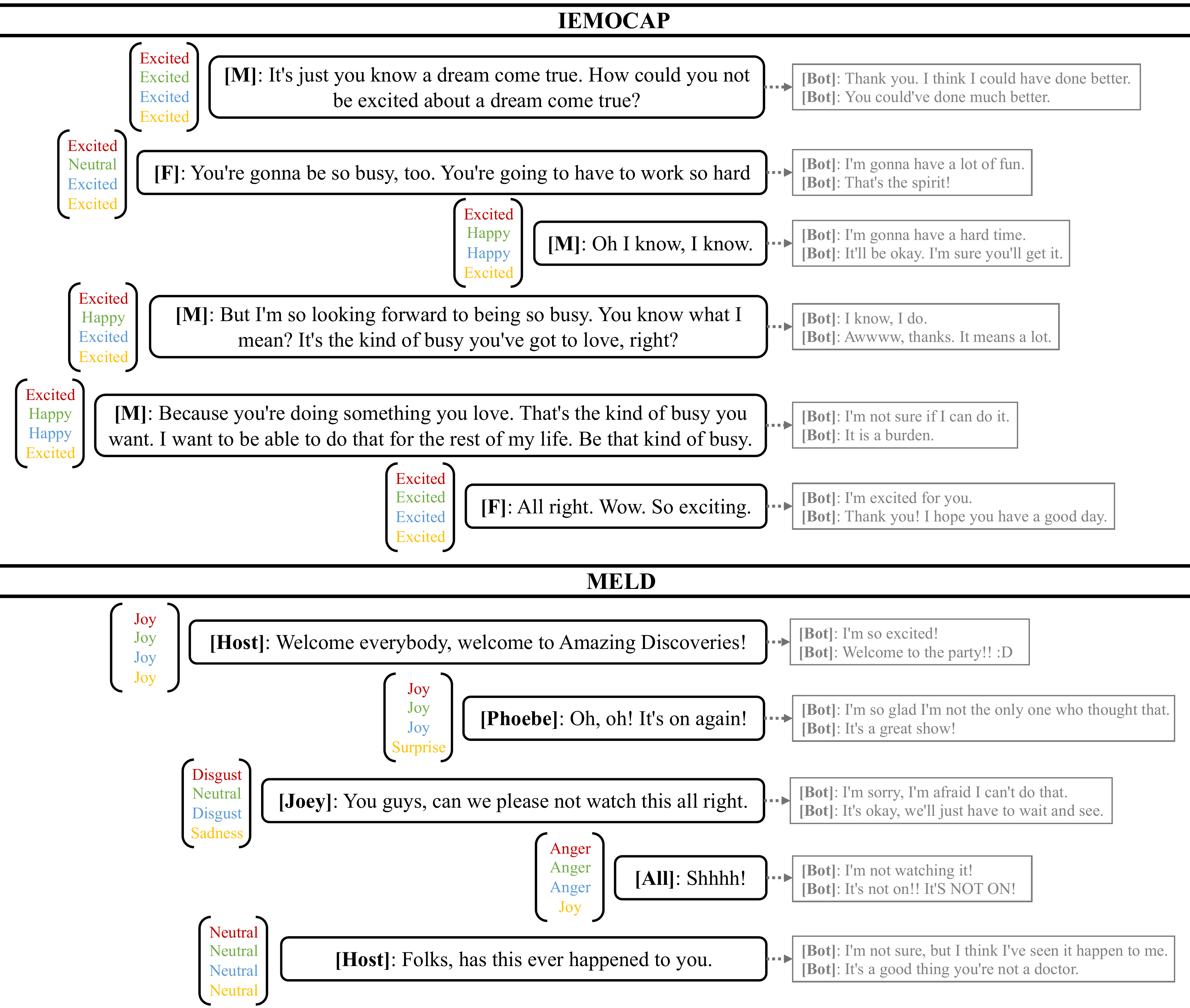}
    \caption{Two cases from IEMOCAP and MELD. In the boxes on the left, from top to bottom, are: labels, predictions from C\&S, C\&S\&PF, and C\&S\&RF settings.}
    \label{fig4}
\end{figure}

We illustrate two examples to promote understanding of our method in Figure~\ref{fig4}.

In the first case from IEMOCAP, the emotions of the two speakers are consistent, which greatly helps to distinguish emotions with similar meanings, such as ``Excited'' and ``Happy'', while pseudo future contexts mislead the model to some extent.

In the second case from MELD, the emotions of multiple speakers are diverse and a high reliance on real future contexts may influence the model's judgement. More consistent supplementary knowledge from pseudo future contexts could help to identify the correct emotions.

\section{Conclusion}
In this paper, we propose a conceptually simple yet effective method of acquiring external homogeneous knowledge by generating pseudo future contexts that are not always available in real-life scenarios. Furthermore, a novel framework named ERCMC is proposed to jointly exploit historical contexts, historical speaker-specific contexts, and pseudo future contexts. Experimental results on four ERC datasets demonstrate the superiority and potential of our method. Further empirical investigations reveal that pseudo future contexts can rival real ones to some extent, especially when the dataset is less context-dependent. Our research could inspire more future works in conversation understanding. In the future, we plan to generate pseudo future contexts in a more controllable way, and extending our method to more tasks.

%
%
%
%
%
%
\bibliographystyle{splncs04}
\bibliography{mybibliography}
%




\end{document}